\documentclass[review]{elsarticle}

\usepackage{lineno,hyperref}
\usepackage{amsmath}
\usepackage[noend]{algpseudocode}
\usepackage{algorithmicx,algorithm}
\usepackage{multirow}
\usepackage{bm}
\usepackage{diagbox}
\modulolinenumbers[5]

\journal{Journal of \LaTeX\ Templates}

\begin{document}

\begin{frontmatter}

\title{LRF-Net: Learning Local Reference Frames for 3D Local Shape Description and Matching}

\author[mymainaddress]{Angfan Zhu}
\ead{zhuangfan@hust.edu.cn}
\author[mysecondaryaddress]{Jiaqi Yang}
\ead{jqyang@nwpu.edu.cn}
\author[mymainaddress]{Weiyue Zhao}
\ead{zhaoweiyue@hust.edu.cn}
\author[mymainaddress]{Zhiguo Cao\corref{mycorrespondingauthor}}
\cortext[mycorrespondingauthor]{Corresponding author}
\ead{zgcao@hust.edu.cn}

\address[mymainaddress]{National Key Laboratory of Science and Technology on Multi-Spectral Information Processing, School of Artificial Intelligence and Automation, Huazhong University of Science and Technology, Wuhan 430074, China.}
\address[mysecondaryaddress]{School of Computer Science, Northwestern Polytechnical University and the National Engineering Laboratory for Integrated aero-Space-Ground-Ocean Big Data Application Technology, Xi'an 710129, China.}

\begin{abstract}
The local reference frame (LRF) acts as a critical role in 3D local shape description and matching. However, most of existing LRFs are hand-crafted and suffer from limited repeatability and robustness. This paper presents the first attempt to learn an LRF via a Siamese network that needs weak supervision only. In particular, we argue that each neighboring point in the local surface gives a unique contribution to LRF construction and measure such contributions via learned weights. Extensive analysis and comparative experiments on three public datasets addressing different application scenarios have demonstrated that LRF-Net is more repeatable and robust than several state-of-the-art LRF methods (LRF-Net is only trained on one dataset). In addition, LRF-Net can significantly boost the local shape description and 6-DoF pose estimation performance when matching 3D point clouds.
\end{abstract}


\end{frontmatter}


\section{Introduction}
The local reference frame (LRF) is a canonical coordinate system established in the 3D local surface, which is a useful geometric cue for 3D point clouds. LRF possesses two intriguing traits. One is that rotation invariance can be achieved via LRF if the local surface is transformed with respect to the LRF~\cite{rops2013l}. The other is that useful geometric information can be mined with LRF~\cite{petrelli2011repeatability}. These make LRF popular in many geometric relevant tasks, especially for local shape description and six-degree-of-free (6-DoF) pose estimation.

For local shape description, two corresponding local surfaces can be converted into the same pose and full 3D geometric information can be employed, which is beneficial to improving the performance of local descriptors. Some hand-crafted local shape descriptors, e.g., signature of histograms of orientations (SHOT) \cite{shot2010} and signature of rotational projection statistics (RoPS) \cite{rops2013l}, estimate an LRF from the local surface and then translate local geometric information with respect to the estimated LRF into distinctive and rotation-invariant feature representations. Some learned local descriptors, e.g., \cite{gojcic2019perfect} and \cite{spezialetti2019learning}, leverage LRFs to overcome the limitation of geometric deep learning networks of being sensitive to rotations. Therefore, LRF is critical for both traditional and learned local shape descriptors. For 6-DoF pose estimation, an LRF can significantly improves its efficiency. Traditional 6-DoF pose estimation is usually performed via RANSAC \cite{derpanis2010overview}, which randomly selects inlier correspondences from an initial correspondence pool to for pose prediction. Such random sampling method is neither reliable nor computational efficient \cite{deng20193d}. By contrast, we can directly predict an initial pose via two corresponding LRFs, reducing the computational complexity from $O(n^3)$ to $O(n)$.
\begin{figure}[t]
	\centering{\includegraphics[width=1\linewidth]{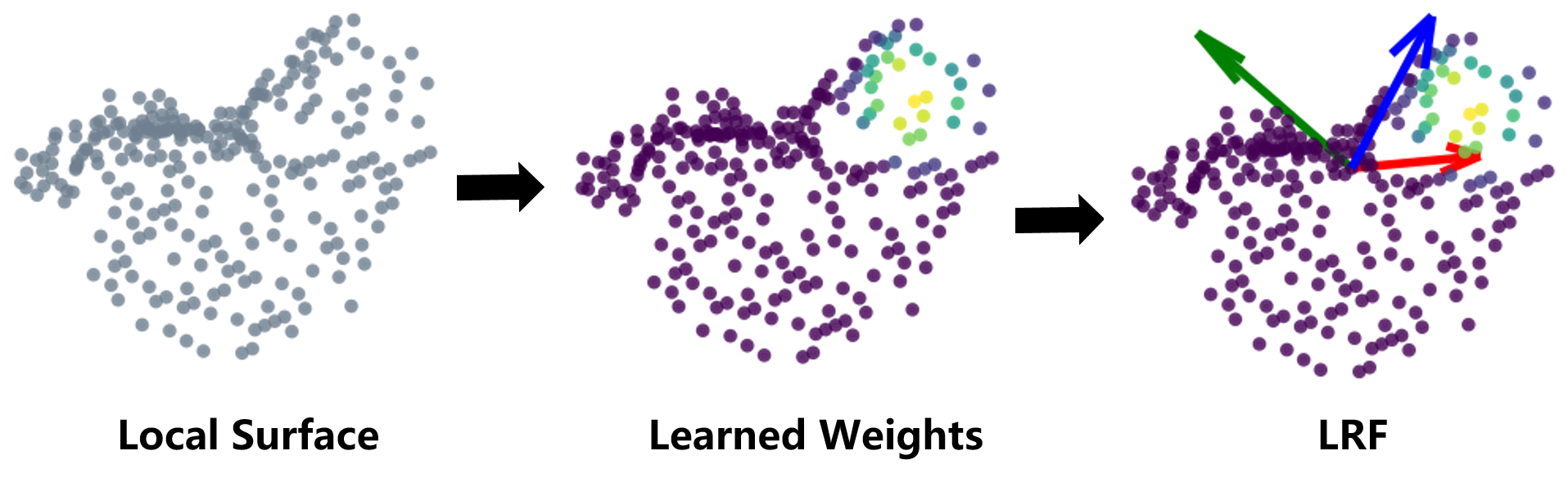}}
	\caption{LRF-Net first assigns learned weights to points in a local surface and then using these weights to estimate a repeatable and robust LRF.}
	\label{fig_fig1}	
\end{figure}

\begin{figure*}[t]
	\centering{\includegraphics[width=1\linewidth]{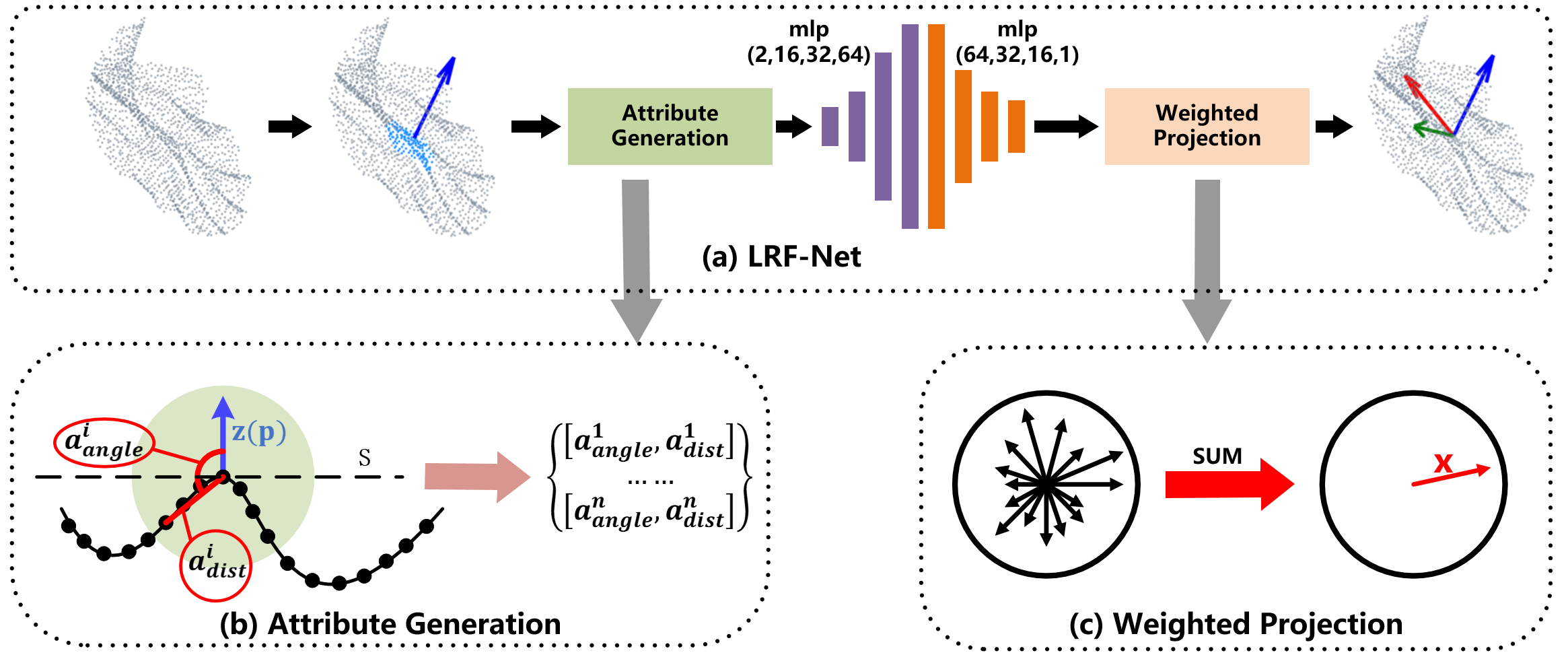}}
	\caption{The architecture of LRF-Net. The input to LRF-Net is a local surface and we calculate its normal as the $z$-axis of the LRF. Then, the local surface is converted to a set of rotation variant attributes. Next, a projection weight for every point is computed with mlp. At last, $x$-axis is calculated by the weighted vector-sum of all the projection vectors and the $y$-axis is calculated by the cross product between $z$-axis and $x$-axis. The LRF is formed as the combination of $x$-axis, $y$-axis and $z$-axis. }
	\label{fig_LRFNet}	
\end{figure*}

The desirable properties for LRF are twofold \cite{shot2010}. The first one is the invariance to rigid transformation (e.g., translations and rotations). The second one is the robustness to common disturbances (e.g., noise, clutter, occlusion and varying mesh resolutions). To achieve these goals, many LRF methods have been proposed in the past decade and they can be categorized into two classes \cite{yang2018toward}: covariance analysis (CA) \cite{mian2010repeatability,shot2010} or point spatial distributions (PSD)-based \cite{petrelli2011repeatability,petrelli2012repeatable,yang2017toldi}. CA-based LRFs are based on the computation of eigenvectors of a covariance matrix calculated either for the points or triangles in the local surface. PSD-based LRFs usually calculate estimate axes successively, where the main efforts are put on the determination of the $x$-axis \cite{yang2018toward}. However, most CA-based LRFs still suffer from sign ambiguity, and PSD-based LRFs show limited robustness to high levels of noise and variations of mesh resolution \cite{petrelli2012repeatable}. Methods in both classes usually apply a weighted strategy to improve their repeatability performance. However, their weights are determined heuristically, and the repeatability performance in challenging 3D matching cases cannot be guaranteed.

Motivated by existing considerations, we propose a learned approach toward LRF estimation (named LRF-Net), which considers the contribution of all neighboring points (Fig.~\ref{fig_fig1}). Our key insight is that each neighboring point in the local surface gives a unique contribution to LRF construction, which can be quantitatively represented by assigning weights to these points. Given a local surface centered at a keypoint, we first resort to the normal of the keypoint computed within a subset of the radius neighbors for the calculation of its $z$-axis. Its repeatability has been confirmed in \cite{petrelli2011repeatability}. Compared with $z$-axis, estimating the $x$-axis is more challenging, due to noise, clutter, and occlusion. By collecting angle and distance attributes within a local neighborhood, we can formulate the estimation of $x$-axis as a weighted prediction problem with respect to these geometric attributes. Note that, we choose these invariant geometric attributes instead of raw points as input to our LRF-Net. We have conduct an experiment to confirm that such attributes can achieve rotation invariance and boost network performance. Unlike previous CA-based and PSD-based approaches, such learned strategy of determining weights is shown to be invariant to rigid transformation and robust to noise, clutter, occlusion and varying mesh resolutions. Our network can be trained in a weakly supervised manner. Specifically, it needs the corresponding relationships between local patches only, instead of ground-truth LRFs and/or exact pose variation information between patches. We have conducted a set of experiments on three public datasets to comprehensively evaluate the proposed LRF-Net. Extensive analysis and comparative experiments on three public datasets addressing different application scenarios have demonstrated that LRF-Net is more repeatable and robust than several state-of-the-art LRF methods (LRF-Net is only trained on one dataset). In addition, LRF-Net can significantly boost the local shape description and 6-DoF pose estimation performance when matching 3D point clouds. The major contributions of this paper are summarized as follows:
\begin{itemize}
	\item LRF-Net, based on a Siamese network that needs weak supervision only, is proposed that achieves the state-of-the-art repeatability performance under the impacts of noise, varying mesh resolutions, clutter and occlusion. To the best of our knowledge, we are the first to concentrate on designing LRF for local surfaces with deep learning.  
	\item LRF-Net can significantly boost the performance of local shape description and 6-DoF pose estimation.
\end{itemize}

The rest of this paper is organized as follows. Section 2 presents a detailed description of our proposed LRF-Net. Section 3 presents the experimental evaluation of LRF-Net on three public datasets with comparisons with several state-of-the-art methods. Several concluding remarks are drawn in Section 4.

\section{Related work}
Various methods for building  LRFs have been proposed in the literature. Most of them can be categorized  into two classes: CA-based methods and PSD-based methods. Given a local surface with a spherical support of radius $r$ centered at the keypoint $p$, they compute a $3\times 3$ matrix as its LRF.

\subsection{CA-based LRF methods}
Most CA-based methods are based on the eigenvectors of the covariance matrix, which is usually generated by the points or triangles in the support region.

\textbf{Mian et al.} \cite{mian2010repeatability}: This method directly calculates the unit vectors of the LRF via computing covariance analysis on the radius neighbors of the keypoint, the three eigenvectors of the covariance matrix are defined as the $x$,$y$,$z$-axis,respectively. While the eigenvectors of the covariance matrix define the principal direction of the local surface, their sign is still ambiguous \cite{petrelli2012repeatable}. Mian et al. disambiguates the sign of $z$-axis through the inner product between $\textbf{n}(p)$ (normal of keypoint $p$) and two possible vector, i.e., $\textbf{z}(p)$ and $\textbf{-z}(p)$, where $\textbf{z}(p)$ denotes the $z$-axis. However, the rest axes are still suffer from sign ambiguity.

\textbf{SHOT} \cite{shot2010}: This method leverages a weighted covariance matrix for the computation of LRF, which assigns smaller weights to more distant points. The weighted covariance matrix is calculated as follows:
\begin{equation}
\textbf{C}_{shot} = \frac{1}{\sum_{q\in N(p)}w_q}\sum_{q\in N(p)}w_q(q-p)(q-p)^T
\label{eq_shot}
\end{equation}
where $w_q=R-||q-p||$. $R$ denotes the support radius and $||\cdot||$ represents $L_2$ norm. This weighted strategy improves the repeatability in present of clutter under 3D object recognition scenarios. To eliminate all sign ambiguities of the LRF axes, a technique which is similar to \cite{bro2008resolving} is applied to the eigenvectors of the weighted covariance matrix. Specifically, the sign of a eigenvector is reoriented to coherent with the majority of the vectors. Such technique is used on the $x$-axis and $z$-axis. The rest $y$-axis is calculated by the cross-product operation between the $z$-axis and the $x$-axis.

\textbf{RoPS} \cite{rops2013l}:  This method does not only calculate one covariance matrix for the local surface, it aggregates multiple covariance matrices computed for every single triangle of the local surface into a comprehensive one to enhance the robustness. Such method needs mesh representation of the 3D local surface. For a triangle $\tau \in \psi (p)$, its covariance matrix is calculated as:
\begin{equation}
C_{\tau}=\frac{1}{12}\sum_{i=1}^{3}\sum_{j=1}^{3}(q_i^{\tau}-p)(q_j^{\tau}-p)^T+\frac{1}{12}\sum_{i=1}^{3}(q_i^{\tau}-p)(q_i^{\tau}-p)^T
\label{eq_triangle}
\end{equation}
where $q_1^{\tau}$, $q_2^{\tau}$ and $q_3^{\tau}$ denote the three vertices of $\tau$. And then, the comprehensive covariance matrix is calculated as:
\begin{equation}
C_{rops}=\sum_{\tau \in \psi (p)} w_1w_2C{\tau}
\label{eq_rops}
\end{equation}
$w_1$ and $w_2$ are defined as: 
\begin{equation}
w_1=\frac{|(q_2^{\tau}-q_1^{\tau})\times (q_3^{\tau}-q_1^{\tau})|}{\sum_{\tau \in \psi (p)}|(q_2^{\tau}-q_1^{\tau})\times (q_3^{\tau}-q_1^{\tau})|}
\label{eq_rops_w1}
\end{equation}
\begin{equation}
w_2=(R-|p-\frac{q_1^{\tau}+q_2^{\tau}+q_3^{\tau}}{3}|)^2
\label{eq_rops_w2}
\end{equation}
where $w_1$ alleviates the impact of mesh resolution variations and $w_2$ improves the robustness performance to clutter and occlusion \cite{yang2018toward}. Based on the eigenvalue decomposition of $C_{rops}$, the three axes of LRF can be calculated.

As for disambiguating the sign, $x$-axis and $z$-axis (only take $x$-axis as an example) are further adjusted via $\textbf{x}(p) = \textbf{x}(p)\cdot sign(h)$, where $\textbf{x}(p)$ denotes the $x$-axis and $h$ is a signum function, which is defined as:
\begin{equation}
h=\sum_{\tau \in \psi (p)}w_1w_2(\frac{1}{6}\sum_{i=1}^{3}(q_i^{\tau}-p)\cdot \bf{x}(p))
\label{eq_rops_sign}
\end{equation}

Once the $x$-axis and $z$-axis are determined, the $y$-axis can be calculated via the cross-product between them.

\subsection{PSD-based LRF methods}
As for PSD-based LRF methods, they calculate three axes of the LRF successively.

\textbf{PS} \cite{chua1997point}: This method puts a sphere of radius $r$ on the keypoint $p$ and gain a contour at the intersection of the local surface. The point with the biggest signed projection distance to the tangent plane of the keypoint was selected to compute the $x$-axis, while the tangent plane is determined by $z$-axis, which is directly performed by the normal of the keypoint. The $y$-axis is calculated via the cross-product operation.

\textbf{Board} \cite{petrelli2011repeatability}: This method collects a small subset of the local surface for the estimation of the $z$-axis, which has achieved a robust performance to occlusion. The $x$-axis is calculated by the points lying in the border region. They choose the point lying in the border region with the biggest deviation angle between its normal and the $z$-axis as the calculation of $x$-axis. And the $y$-axis is computed by the cross-product operation between $z$-axis and $x$-axis. 

\textbf{SD} \cite{petrelli2012repeatable}: This method is a modified version of Board \cite{petrelli2011repeatability}. They make improvement to the repeatability of the LRF via employing the point with largest local depth instead of deviation angle in SD \cite{petrelli2012repeatable}. They achieve a more repeatable performance than Board on 3D registration and recognition data. However, both of them show a weak performance on the robustness to the large scale noise.

\textbf{TOLDI} \cite{yang2017toldi}: This method resorts to the normal of the keypoint which is calculated by a subset of the radius neighbors for the estimation of its $z$-axis. Then, the tangent plane of the keypoint with respect to $z$-axis is determined and all radius neighbors of the keypoint are projected on the tangent plane. A weighted strategy is employed to each projection vector to calculate the $x$-axis, which is defined as:
\begin{equation}
w_{i1} = (r-||p-q_i||)^2
\label{eq_w1}
\end{equation}
\begin{equation}
w_{i2}=(\bf{pq}_{i}\cdot \bf{z}(p))^2
\label{eq_w2}
\end{equation}
where $p$ donates the keypoint and $q_i$ is one of its radius neighbors within support radius $r$. $w_{i1}$ is a weight related to the distance from $p$ to $q_i$, which is designed to improve the robustness of the LRF to clutter, occlusion and incomplete border regions \cite{yang2017toldi}. $w_{i2}$ is a weight related to the local depth which is designed to provide high repeatability on flat regions \cite{yang2017toldi}. And the $x$-axis is calculated as:
\begin{equation}
{\bf{x(p)}}=\sum_{i=1}^{k} w_{i1}w_{i2}{\bf{v}}_{i}/\left\|\sum_{i=1}^{k}w_{i1}w_{i2}{\bf{v}}_{i}\right\|
\label{eq_toldi}
\end{equation}
where $k$ is the count of radius neighbors of keypoint $p$ and $\textbf{v}_i$ denotes one of the projection vectors. And the $y$-axis is computed by the cross-product between $z$-axis and $x$-axis.

\textbf{GFrames} \cite{melzi2019gframes}: This method works straight away with mesh triangles. The $z$-axis is calculated as the normal on the point and the $x$-axis is defined as:
\begin{equation}
{\bf{x(p)}} = \frac{1}{\sum_{t_j \in N_{R}(p)}A(t_j)}\sum_{t_j \in N_{R}(p)}A(t_j)\nabla f(t_j)
\label{eq_GFrame}
\end{equation}
where $t_j$ is a mesh triangle, $N_{R}(p)$ denotes the set of triangles within distance $R$ from $p$, $A(t_j)$ represents the area of the $t_j$ triangle and $f$ is a user-defined scalar function.

The final $x$-axis is calculated by projecting $\bf{x(p)}$ on the tangent plane and normalized into a unit vector. The $y$-axis is computed by the usual cross-product. Many $f$ functions have been tested in \cite{melzi2019gframes}, such as the mean curvature, the Gaussian curvature, and the sum of total Euclidean distances, displaying its great robustness and repeatability. Own to its flexibility, GFrames is also suited for non-rigid transformations.

\section{Method}
This section represents the details of our proposed LRF-Net for 3D local surface. We first introduce the technique approach for calculating the three axes for an LRF and then describes a weakly supervised approach for training LRF-Net. 

\subsection{A Learned LRF Proposal}	
The whole architecture of LRF-Net in shown in Fig.~\ref{fig_LRFNet}(a). LRF-Net predicts  the direction of three axes successively. For a local surface, we first estimate its $z$-axis via its normal vector computed over a small subset of the local point set. Then, unique weights are learned for each point in the local surface. The $x$-axis is calculated by integrating projection vectors with learned weights using a vector-sum operation. At last, the $y$-axis is calculated by the cross-product operation between $z$-axis and $x$-axis.
\\\\\noindent\textbf{LRF definition:}	Given a local surface $\bf{Q}$ centered at keypoint $\bf{p}$, the LRF at $\bf{p}$ (denoted by $\bf{L_p}$) can be represented as :
\begin{equation}
	\bf{L_p}=[\bf{x(p)}, \bf{z(p)}\times \bf{x(p)}, \bf{z(p)}],\label{eq_lrf}
\end{equation}
where $\bf{x(p)}$, $\bf{y(p)}$, and $\bf{z(p)}$ denote the $x$-axis, $y$-axis, and $z$-axis of $\bf{L_p}$, respectively. As three axes are orthogonal, the estimation of LRF therefore contains two parts: estimation of the $z$-axis and the $x$-axis. 

A naive way to learn an LRF for the local surface is to train a network that directly regresses the axes. The premise is that ground-truth LRFs are labeled for local surfaces. Unfortunately, the network trained in this manner meets two difficulties. The first one is that the definition of ground-truth LRFs for local surfaces remain an open issue in the community~\cite{yang2018toward}. The second one, which is more important,  is that the orthogonality of three axes cannot be guaranteed. We suggest estimating $z$-axis and $x$-axis independently.
\\\\\noindent\textbf{z-axis:} As for $z$-axis, we take the normal of the keypoint as the $z$-axis., which has been confirmed \cite{petrelli2011repeatability} to be quite repeatable. To resist the impact fo clutter and occlusion, we collect a small subset of the local surface to calculate the normal. For more details, readers are referred to \cite{yang2017toldi}. 
\\\\\noindent\textbf{x-axis:} Once the $z$-axis is determined, the remaining task is to compute the $x$-axis. Compared with $z$-axis, $x$-axis is  more challenging due to noise, clutter, and occlusion~\cite{yang2018toward}.	
We argue that each neighboring point in the local surface gives a unique contribution to LRF construction. Hence, we predict a weight for each neighboring point and leverage all neighboring points with learned weights for $x$-axis prediction. The main steps are as follows.
\begin{figure}[t]
	\centering{\includegraphics[width=1\linewidth]{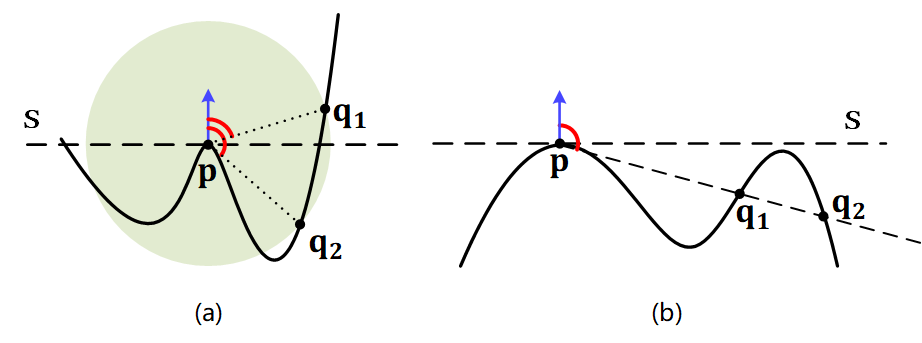}}
	\caption{An illustration of information complementary inherent to the two attributes in LRF-Net. The two radius neighbors ${\bf q}_1$ and ${\bf q}_2$ of the keypoint $\bf p$ in (a) and (b) have different spatial locations. In (a), the two radius neighbors with the same distance value are distinguished by the surface variation angle attribute. In (b), their surface variation angle attribute values are similar, while can be distinguished by the distance attribute.}
	\label{fig_insight}	
\end{figure}
First, to make the estimate LRF invariant to rigid transformation, our network consumes with invariant geometric attributes, rather than point coordinates. In particular, two attributes, i.e., relative distance $a_{dist}$ and surface variation angle $a_{angle}$ are used in LRF-Net as illustrated in Fig.~\ref{fig_LRFNet}(b). For a neighbor $\bf{q}_i$ of $\bf{p}$, the two attributes of $\bf{q}_i$ are computed as:
\begin{equation}
	\begin{cases}
		a_{dist}^i = \left\|\bf{pq}_{i}\right\|/r\\
		a_{angle}^i = \cos(\bf{z(p)}, \bf{pq}_{i})	
	\end{cases},
	\label{eq_attrib}
\end{equation}
where $ \left\|\cdot\right\| $ is the $L_2$ norm and $r$ represents the support radius of the local surface. The range of $a_{angle}$ and $a_{dist}$ are $[-1, 1]$ and $[0, 1]$, respectively. Thus, every radius neighboring point represented by two attributes that will be encoded to a weight value via LRF-Net later. 	
The employed two attributes in LRF-Net have two merits at least. First, the unique spatial information of a radius neighboring point in the local surface can be well represented, as shown in Fig.~\ref{fig_insight}. Both attributes are complementary to each other. Second, the two attributes are calculated with respect to the keypoint, which are rotation invariant. It makes the learned weights rotation invariant as well.
\begin{figure}[t]
	\centering{\includegraphics[width=0.5\linewidth]{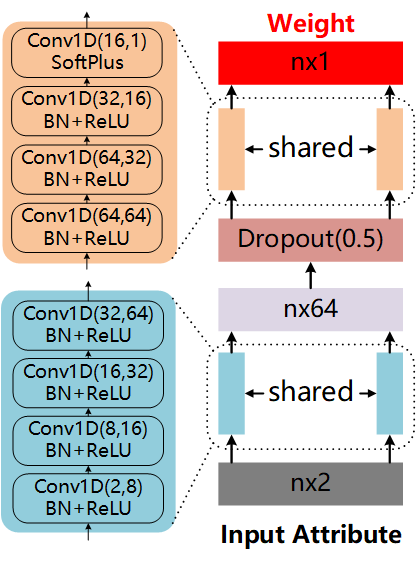}}
	\caption{Parameters of our LRF-Net.}
	\label{fig_net}	
\end{figure}
Second, with geometric attributes being the input, we use a U-Net with multilayer perceptions (MLP) layers only to predict weights for neighboring points. The details of the network are illustrated in  Fig.~\ref{fig_net}. The network is very simple, however, is sufficient to predict stable and informative weights for neighboring points (as will be verified in the experiments).

Third, because $x$-axis is orthogonal to $z$-axis, we project each neighbor $\bf{q}_{i}$ on the tangent plane $\bf S$ of the $z$-axis and compute a projection vector for $\bf{q}_{i}$ as:
\begin{equation}
	{\bf{v}}_{i}={\bf{pq}}_{i}-({\bf{pq}}_{i}\cdot {\bf{z(p)}})\cdot \bf{z(p)}.
	\label{eq_vector_x}
\end{equation}
We integrate all weighted projection vectors in a weighted vector-sum manner: 
\begin{equation}
	{\bf{x(p)}}=\sum_{i=1}^{n}w_i{\bf{v}}_{i}/\left\|\sum_{i=1}^{n}w_i{\bf{v}}_{i}\right\|,
	\label{eq_axis_x}
\end{equation}
where $n$ denotes the total number of radius neighbors of keypoint $\bf{p}$ and $w_{i}$ is a learned weight by LRF-Net. Another way for determining the $x$-axis, based on these weights, is choosing the vector with the maximum weight, as in many PSD-based LRFs~\cite{petrelli2011repeatability,petrelli2012repeatable}. However, it fails to leverage all neighboring information and we will shown that it is inferior to the vector-sum operation in the experiments. 
\\\\\noindent\textbf{y-axis:} Based on the calculated $z$-axis and $x$-axis, the $y$-axis can be computed by the cross-product between them.

\subsection{Weakly Supervised Training Scheme}
Our training data are constituted by a series of corresponding local surface patches. The corresponding relationship is obtained based on the ground-truth rigid transformation of two whole point clouds. In particular, LRF-Net needs the corresponding relationships between local surface patches only, rather than ground-truth LRFs and/or exact pose variation information between patches. Therefore, our network can be trained in a weakly supervised manner. 

We train our LRF-Net with two streams in a Siamese fashion where each stream independently predicts an LRF for a local surface. Specifically, two streams take the local surfaces of keypoints $\bf{p}_{m}$ and $\bf{p}_{s}$ as inputs, respectively. Here, $\bf{p}_{m}$ and $\bf{p}_{s}$ are two corresponding keypoints sampled from the model and scene point cloud. Both streams share the same architecture and underlying weights. We use the predicted LRFs ${\bf L}_m$ and ${\bf L}_s$  by two stream to transform the local surfaces ${\bf Q}_m$ and ${\bf Q}_s$ to the coordinate system of the two LRFs. Then, we calculate the Chamfer Distance \cite{deng2018ppf} between two transformed local surfaces as the loss function to train LRF-Net:
\begin{equation}
	Loss = d_{cham}({\bf L}_m\cdot{\bf Q}_m, {\bf L}_s\cdot{\bf Q}_s),
	\label{eq_loss}
\end{equation}
where
\begin{equation}
	\begin{split}
		d_{cham}(X,\hat{X})= \min \left\{ \frac{1}{|X|}\sum_{x\in X}\min_{\hat{x}\in \hat{X}}||x - \hat{x}||, \frac{1}{|\hat{X}|}\sum_{\hat{x}\in \hat{X}}\min_{x\in X}||x-\hat{x}|| \right\}.
		\label{eq_cham}
	\end{split}
\end{equation}

Remarkably, our opinion is that it is difficult to define a ``good'' LRF for a single local surface. For 3D shape matching, LRFs that can align the poses of two local surface patches are judged as repeatable. This motivates us to consider two local patches simultaneously and  employ the Chamfer Distance to train the network.

\section{Experiments}
In this section, we first evaluate the repeatability performance of our LRF-Net on three standard datasets, including the Bologna retrieval (BR) dataset \cite{tombari2013performance}, the UWA 3D modeling (UWA3M) dataset \cite{mian2006novel}, and the UWA object recognition (UWAOR) dataset \cite{mian2006three}, together with a comparison with other state-of-the-art LRFs. Second, we apply our LRF-Net perform local shape description and 6-DoF pose estimation to verify the practicability of our method. 
Third, analysis experiments are conducted to improve the explainability of the proposed LRF-Net.
\subsection{Experimental Setup}
The details of our experiments including the description of datasets and the illustration for all compared methods are introduced before evaluation. The experiments were conducted  on a Windows Server with an Intel Xeon E5-2640 2.39 GHz CPU and 96 GB of RAM. We train our LRF-Net using a batch size of 512 local surface pairs and leverage the ADAM optimizer with an initial learning rate of 1e-4, which decays $5\%$ every epoch. Each sampled local surface contains 256 points. The max epoch count is set to 20.

\subsubsection{Datasets}
\begin{figure}
	\centering{\includegraphics[scale=0.6]{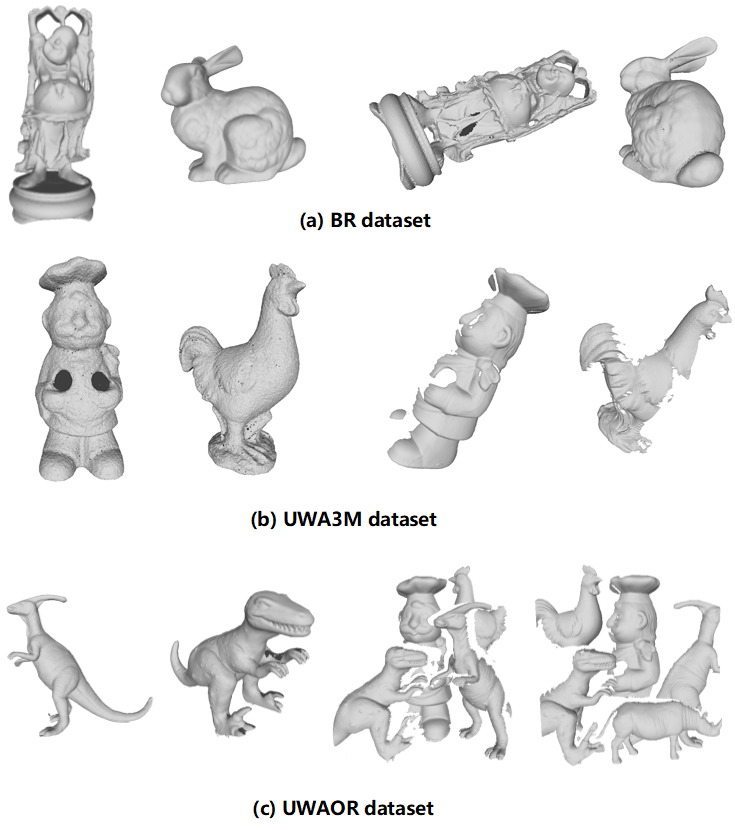}}
	\caption{Two exemplar models and scenes without noise (shown from left to right) respectively taken from the BR, UWA3M, and UWAOR datasets.}
	\label{fig_dataset}
\end{figure}

Our experimental datasets includes three standard datasets with different application scenarios. The variety among these public 3D datasets definitely helps us to evaluate the performance of our method in a comprehensive manner. Fig.~\ref{fig_dataset} displays two exemplar models and scenes without noise in each dataset. The main properties of these datasets are summarized in Table~\ref{tabel_dataset}.

These dataset are also injected with five levels of Gaussian noise (i.e., from 0.1 mr to 0.5 mr Gaussian noise) and four levels of mesh decimation (i.e., $\frac{1}{2}$, $\frac{1}{4}$, $\frac{1}{8}$ and $\frac{1}{16}$ of original mesh resolution). Here, the unit mr denotes mesh resolution. Remarkably, \textit{the noise-free BR dataset is used to train our LRF-Net, the rest noisy data in the BR dataset and data in the UWA3M dataset and the UWAOR dataset are used for testing}. 
\begin{table}[t]
	\scriptsize
	\renewcommand{\arraystretch}{1.0}
	\caption{Experimental datasets and inherited properties}
	\label{tabel_dataset}
	\centering
	\begin{tabular}{c|ccc}
		\hline
		\textbf{Dateset}                                                      & BR                                                        & UWA3M                                                                                  & UWAOR                                                            \\ \hline
		\textbf{Scenario}                                                     & Retrieval                                                 & Registration                                                                           & Recogntion                                                       \\ \hline
		\textbf{Challenge}                                                    & \begin{tabular}[c]{@{}c@{}}Gaussian \\ noise\end{tabular} & \begin{tabular}[c]{@{}c@{}}holes, missing \\ region, and \\ self-occlusion\end{tabular} & \begin{tabular}[c]{@{}c@{}}clutter \\ and occlusion\end{tabular} \\ \hline
		\textbf{\# Models}                                                    & 6                                                         & 4                                                                                      & 5                                                                \\ \hline
		\textbf{\# Scenes}                                                    & 18                                                        & 75                                                                                     & 50                                                               \\ \hline
		\textbf{\begin{tabular}[c]{@{}c@{}}\# Matching  Pairs\end{tabular}} & 18                                                        & 75                                                                                     & 188                                                              \\ \hline
	\end{tabular}
\end{table}

\subsubsection{Compared Methods}
We compare our LRF-Net with several existing LRF methods for a through evaluation. Specifically, the compared methods are proposed by Mian et al.~\cite{mian2010repeatability}, Tombari et al.~\cite{shot2010}, Petrelli et al.~\cite{petrelli2012repeatable}, Guo et al.~\cite{rops2013l} and Yang et al.~\cite{yang2017toldi}, respectively. We dub them as \textit{Mian}, \textit{Tombari}, \textit{Petrelli}, \textit{Guo}, and \textit{Yang}, respectively. To compare fairly,  we keep the support radius of all the LRFs as 15 mr. The properties of these LRFs are shown in Table~\ref{tabel_param}. 

To evaluate the local shape description performance of our method, we replace the LRF in four LRF-based descriptors (i.e., snapshots \cite{malassiotis2007snapshots}, SHOT \cite{shot2010}, RoPS \cite{rops2013l} and TOLDI \cite{yang2017toldi}) and assess the performance variations. To measure the 6-DoF pose estimation performance of our method, we adapt LRF-Net to the RANSAC pipeline and compare with the original RANSAC~\cite{fischler1981random}.  
\begin{table}[t]
	\scriptsize
	\renewcommand{\arraystretch}{1.1}
	\renewcommand\tabcolsep{4.5pt} 
	\caption{Properties of six LRF methods. H and L respectively represent hand-crafted and learned methods for point weight calculation; P and M respectively denote point cloud and mesh.}
	\label{tabel_param}
	\centering
	\begin{tabular}{c|cccccc}
		\hline
		\textbf{Method}    & \textit{Mian} & \textit{Tombari} & \textit{Guo} & \textit{Petrelli} & \textit{Yang} & \textit{Ours} \\ \hline
		\textbf{Category}  & CA            & CA               & CA           & PSD               & PSD           & PSD           \\ \hline
		
		\textbf{Date type} & P             & P                & M            & P                 & P             & P             \\ \hline
		\textbf{Weight}    & $-$             & H                & H            & H                 & H             & L             \\ \hline
	\end{tabular}
\end{table}

\subsection{Performance Evaluation of LRF-Net}
\subsubsection{Repeatability Performance}
\begin{figure}[t]
	\centering{\includegraphics[width=0.8\linewidth]{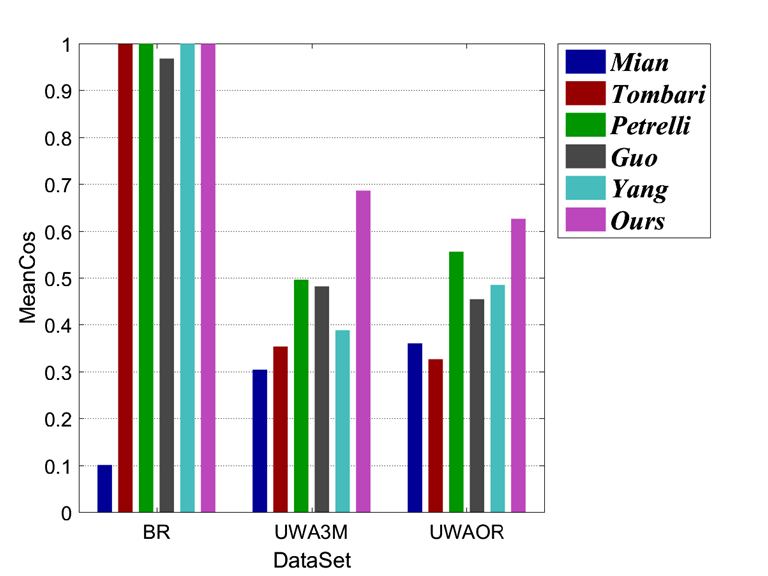}}
	\caption{Repeatability performance of six LRF methods on the BR, UWA3M, and UWAOR datasets.}
	\label{fig_meancos}
\end{figure}

\begin{figure}[t]
	\centering{\includegraphics[width=1\linewidth]{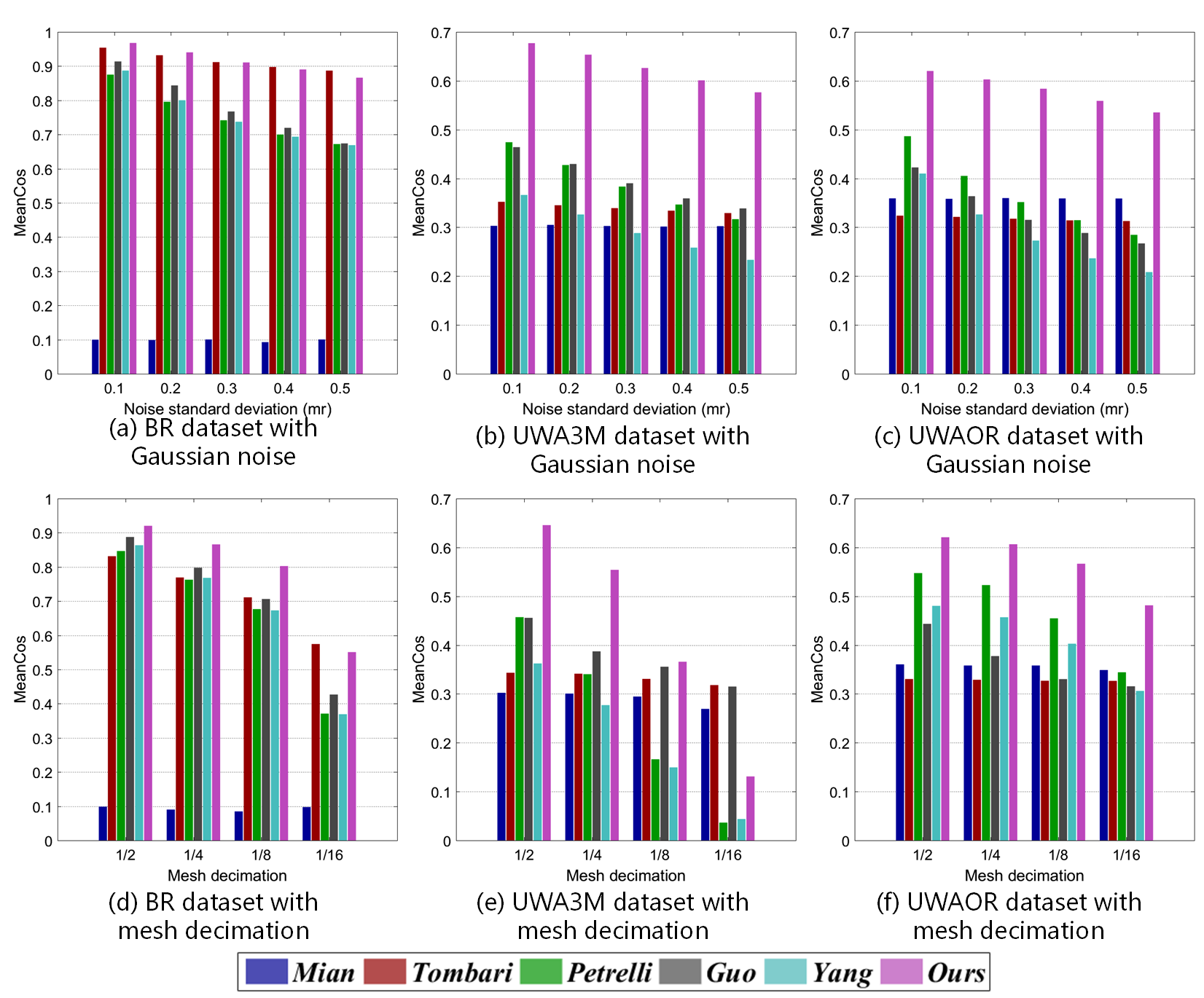}}
	\caption{Robustness performance of six LRF methods on the BR, UWA3M, and UWAOR	datasets with Gaussian noise and mesh decimation.}
	\label{fig_robustness}
\end{figure}
We evaluate the repeatability of all LRFs via the popular $MeanCos$ \cite{shot2010} metric, which measures overall angular error between two LRFs. The $MeanCos$ criterion is computed as:
\begin{equation}
MeanCos(\textbf{L}_m, \textbf{L}_{s}^{'})=\frac{Cos(X)+Cos(Z)}{2}
\label{eq_meancos}
\end{equation}
\begin{equation}
\textbf{L}_{s}^{'} =  \textbf{L}_{s}* \textbf{GT}
\label{eq_LRFtrans}
\end{equation}
where $\textbf{L}_m$ and $\textbf{L}_s$ denote two corresponding LRFs between model and scene. $\textbf{L}_{s}^{'}$ represents the transformed $\textbf{L}_s$, gained via ground truth transformation $\textbf{GT}$. $*$ denotes matrix-product. $Cos(Z)$ represents the cosine of the angle between the $z$-axis of the $\textbf{L}_m$ and the $\textbf{L}_{s}^{'}$, and $Cos(X)$ coincides with the $x$-axis angular error between $\textbf{L}_m$ and the $\textbf{L}_{s}^{'}$. Due to the $y$-axis can always be calculated from the other two axes via cross-product, it is not necessary to be included in $MeanCos$ calculation \cite{petrelli2011repeatability}. In our evaluation, we first randomly select 1000 points from each models and collect the corresponding points in the scenes via ground truth transformation for each model-scene pair. Then, we calculate the LRF for every local surface centered at the selected point in the model and scene. At last, the average $MeanCos$ of the $MeanCos$ value of all the corresponding LRFs between each model-scene pair is calculated as the final result for a dataset. Note that, the $MeanCos$ of two perfectly corresponding LRFs equals to 1. The repeatability results of evaluated LRFs are shown in Fig.~\ref{fig_meancos} and Fig.~\ref{fig_robustness}. Several observations can be made from these figures.

First, as witnessed by Fig.~\ref{fig_meancos}, our LRF together with \textit{Tombari}, \textit{Petrelli}, and \textit{Yang} achieve decent performance on the BR dataset. On the UWA3M and UWAOR datasets, our LRF-Net achieves the best performance. Second, as shown in Fig.~\ref{fig_robustness}(a), LRF-Net and \textit{Tombari} achieve a comparably stable performance on the BR dataset with respect to different levels of Gaussian noise. Fig.~\ref{fig_robustness}(b) and Fig.~\ref{fig_robustness}(c) indicate that LRF-Net achieves the best performance under all levels of Gaussian noise on the UWA3M and UWAOR datasets, surpassing the others by a very significant gap. Note that UWA3M and UWAOR datasets also include nuisances such as clutter, self-occlusion, and occlusion. Third, results in  Fig.~\ref{fig_robustness}(d)-(f) suggest that LRF-Net is the best competitor with  $\frac{1}{2}$, $\frac{1}{4}$, and $\frac{1}{8}$ mesh decimation on all datasets. 

These results clearly demonstrate the strong robustness of our LRF-Net with respect to Gaussian noise, mesh decimation, clutter, and occlusion. The reasons are at least twofold. One is that all points are leveraged to generate the critical $x$-axis, which guarantees the robustness to Gaussian noise and low level mesh decimation. The other is that a LRF-Net can learn stable and informative weights for neighboring points. It can improve the robustness of LRF-Net to common nuisances. 

\subsubsection{Local Shape Description Performance}
We further evaluate our LRF-Net by replacing the LRFs in four LRF-based descriptors (i.e., snapshots, SHOT, RoPS, and TOLDI) with our LRF-Net. Then we compare their descriptor matching performance measured via recall vs. 1-precision curve (RPC) \cite{guo2016comprehensive,shot2010}. The calculation of recall is defined as:
\begin{equation}
recall = \frac{N_{true}}{N_{corr}}
\label{eq_recall}
\end{equation}
where $N_{true}$ denotes the number of correct matches and $N_{corr}$ is the total number of corresponding features. The calculation of 1-precision is defined as:
\begin{equation}
1-precision = \frac{N_{false}}{N_{match}}
\label{eq_prec}
\end{equation}
where $N_{false}$ represents the number of false matches and $N_{match}$ is the total number of matches.

\begin{figure}[t]
	\centering{\includegraphics[width=1\linewidth]{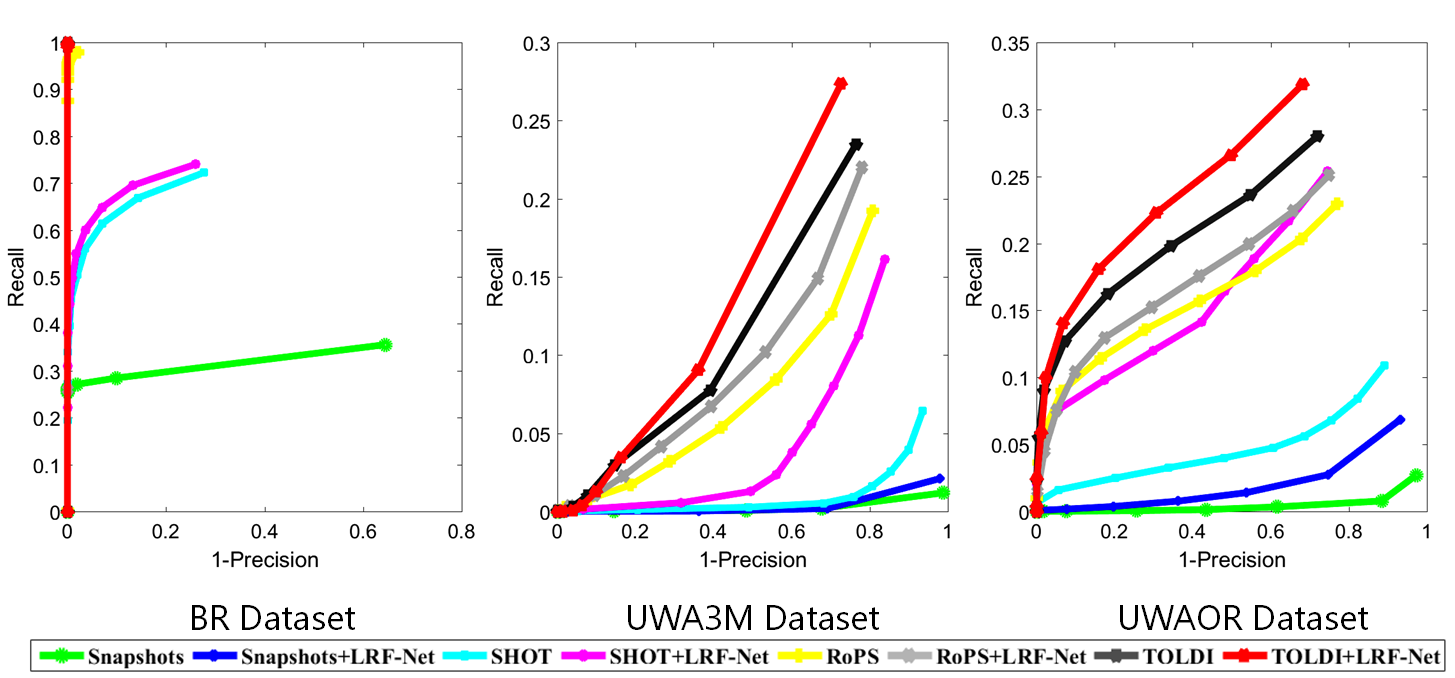}}
	\caption{Local shape description performance of LRF-based descriptors with LRF-Net and their original LRFs on the BR, UWA3M, and UWAOR datasets.}
	\label{fig_RPC}
\end{figure}

Notably, the original LRF methods employed by snapshots, SHOT, RoPS, and TOLDI are \textit{Mian}, \textit{Tombari}, \textit{Guo} \textit{Yang}, respectively. We conduct this experiment on the original BR, UWA3M, and UWAOR datasets. Fig.~\ref{fig_RPC} reports the RPC results of the all tested descriptors.

As witnessed by the figure, all LRF-based descriptors equipped with our LRF-Net outperform their original versions. Specifically, snapshots achieves a dramatic performance improvement with our LRF-Net on the BR dataset; the performance of SHOT also climbs significantly  on the UWA3M and UWAOR datasets with the help of the proposed LRF-Net. Therefore, we can draw a conclusion that LRF plays an important role in local shape description, where a repeatable LRF can effectively improve the description performance of an LRF-based descriptor without changing its feature representation. It also indicates that the proposed LRF-Net can bring positive impacts on a number of existing local shape descriptors.

\subsubsection{6-DoF Pose Estimation Performance}
\begin{figure}
	\centering{\includegraphics[scale=0.4]{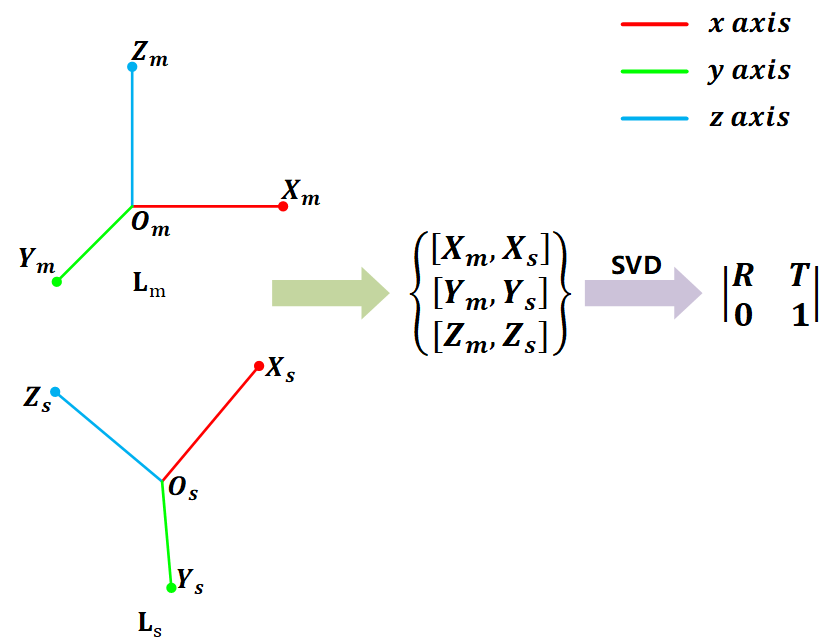}}
	\caption{Illustration of directly calculating an initial pose via a single correspondence. We generate three corresponding point pairs via the centroids and LRFs of the corresponding local surface pair. And the final pose is computed via SVD, which is a inner function in PCL.}
	\label{fig_SVD}
\end{figure}

A general 6-DoF pose estimation process with local descriptors is achieved by correspondence generation and pose estimation from correspondences with potential outliers \cite{derpanis2010overview}. RANSAC is arguablly the de facto 6-DoF pose estimator in many applications. However, a key limitation of RANSAC is that the computational complexity of RANSAC is $O(n^3)$ and estimating a reasonable pose requires a huge number of iterations. With LRFs, a single correspondence is able to generate a 6-DoF pose (shown as Fig.~\ref{fig_SVD}), decreasing the computational complexity from $O(n^3)$ to $O(n)$. Therefore, we apply LRF-Net to 6-DoF pose estimation, following a RANSAC-fashion pipeline. The difference is that we sample one correspondence per iteration. Two criteria, i.e., the rotation error $err_r$ between our predicted rotation $R$ and the ground-truth one $R_{GT}$, and the translation error $err_t$ between the predicted translation vector $T$ and the ground truth one $T_{GT}$ \cite{mian2006novel}, are employed for evaluating the performance of 6-DoF pose estimation. $err_r$ and $err_t$ are defined as:
\begin{equation}
err_r = arccos(\frac{trace(R^{'}-1)}{2})\frac{180}{\pi}
\label{eq_err_r}
\end{equation}

\begin{equation}
err_t = \frac{||T_{GT}-T||}{mr}
\label{eq_err_t}
\end{equation}
where $R^{'}=R_{GT}(R)^{-1}$ and $mr$ denotes the mesh resolution.

The initial feature correspondence set is  generated by first matching TOLDI (equipped with our LRF-Net) descriptors and keeping 100 correspondences with the highest similarity scores. 100 and 1000 iterations are assigned to our method and RANSAC. The average rotation errors and translation errors of the two estimators on three experimental datasets are shown in Table \ref{tabel_pose}.

\begin{table}[t]
	\scriptsize
	\renewcommand{\arraystretch}{1.1}
	\caption{6-DoF Pose estimation performance on three experimental datasets.}
	\label{tabel_pose}
	\centering
	\begin{tabular}{ccccc}
		\hline
		&        & \bf BR     & \bf UWA3M & \bf UWAOR \\ \hline
		\multicolumn{1}{c|}{\multirow{2}{*}{RANSAC}} & $err_t$ & 0.000  & 7.929 & 9.513 \\
		\multicolumn{1}{c|}{}                        & $err_r$ & 0.030 & 0.696 & 0.769 \\ \hline
		\multicolumn{1}{c|}{\multirow{2}{*}{LRF-Net}} & $err_t$ & 0.000  & 6.088 & 4.392 \\
		\multicolumn{1}{c|}{}                        & $err_r$ & 0.024 & 0.608 & 0.405 \\ \hline
	\end{tabular}
\end{table}

Two salient observations can be made from the table. First, both RANSAC and our method manage to achieve accurate pose estimation results on the BR dataset that contains point cloud pairs with large overlapping ratios. However, our method only needs $\frac{1}{10}$ of the iterations required for RANSAC. Second, on more challenging datasets, i.e., UWA3M and UWAOR, our method significantly outperforms RANSAC. This demonstrates that LRF-Net can improve the accuracy and efficiency of RANSAC for 6-DoF pose estimation simultaneously.

\subsection{Analysis Experiments}
\subsubsection{Verifying the Rationality of LRF-Net}
\begin{figure}[t]
	\centering{\includegraphics[width=0.4\linewidth]{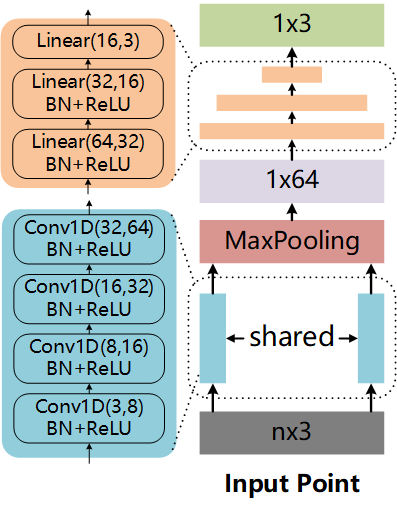}}
	\caption{The architecture of DR.}
	\label{fig_DR}
\end{figure}

To verify the rationality of the main technique components of our LRF-Net, we conduct the following experiments. As mentioned above, our LRF-Net contains two main parts: estimating $z$-axis and $x$-axis. First, in order to verify the choice of normal vector for $z$-axis calculation, we replace the normal vector with the one regressing $z$-axis via a network shown in Fig.~\ref{fig_DR} (dubbed "DR"). Second, to confirm the advantage of our $x$-axis technique, we  perform analysis experiments from three aspects. (1) To prove the advantage of invariant geometric attributes, we replace the invariant geometric attributes with the combination of original points and $z$-axis (i.g., $[{\bf{q}}_i, \bf{z(p)}]$). And then, we calculate the $x$-axis in a weighted vector-sum manner. The former is dubbed "Sum1" and the latter is dubbed "Sum2". (2) In order to verify the choice of weighted vector-sum operation for $x$-axis calculation, we test the approach using the vector with the maximum weight as  the $x$-axis (dubbed ``Max''). (3) To demonstrate that the axes of LRF is not suitable to be directly regressed, we compare our method with the one regressing $x$-axis via a network (DR). There are totally eight different combinations. All of them are tested on BR, UWA3M and UWAOR datasets. The results are shown in Table \ref{analyse_BR}, \ref{analyse_UWA3M}, \ref{analyse_UWAOR}.

Clearly, LRF-Net (Normal + Sum1) achieves the best performance among tested methods. It verifies that learning weights via invariant geometric attributes rather than directly learning axes is more reasonable. In addition, vector-sum is more appropriate for integrating projection vectors with learned weights for LRF-Net.

\begin{table}[t]
	\scriptsize
	\renewcommand{\arraystretch}{1.1}
	\caption{$MeanCos$ performance of eight different combinations on BR Dataset}
	\label{analyse_BR}
	\centering
	\begin{tabular}{|c|c|c|c|c|}
		\hline
		\multicolumn{5}{|c|}{\bf{BR Dataset}}         \\ \hline
		\diagbox{$z$-axis}{$x$-axis} & Sum1       & Sum2       & DR           & Max     \\ \hline
		Normal                           & \bf{0.999} &\bf{0.999}  &  0.775 & 0.720 \\ \hline
		DR                                 & 0.737   & 0.778  & 0.582  & 0.471 \\ \hline
	\end{tabular}
\end{table}

\begin{table}[t]
	\scriptsize
	\renewcommand{\arraystretch}{1.1}
	\caption{$MeanCos$ performance of eight different combinations on UWA3M Dataset}
	\label{analyse_UWA3M}
	\centering
	\begin{tabular}{|c|c|c|c|c|}
		\hline
		\multicolumn{5}{|c|}{\bf{UWA3M Dataset}}         \\ \hline
		\diagbox{$z$-axis}{$x$-axis} & Sum1       & Sum2       & DR           & Max     \\ \hline
		Normal                           & \bf{0.690} & 0.429  &  0.574 & 0.412 \\ \hline
		DR                                 & 0.390   & 0.495  & 0.323  & 0.287 \\ \hline
	\end{tabular}
\end{table}

\begin{table}[t]
	\scriptsize
	\renewcommand{\arraystretch}{1.1}
	\caption{$MeanCos$ performance of eight different combinations on UWAOR Dataset}
	\label{analyse_UWAOR}
	\centering
	\begin{tabular}{|c|c|c|c|c|}
		\hline
		\multicolumn{5}{|c|}{\bf{UWAOR Dataset}}         \\ \hline
		\diagbox{$z$-axis}{$x$-axis} & Sum1       & Sum2       & DR           & Max     \\ \hline
		Normal                           & \bf{0.624} & 0.432  &  0.528 & 0.380 \\ \hline
		DR                                 & 0.408   & 0.490  & 0.467  & 0.366 \\ \hline
	\end{tabular}
\end{table}

\subsubsection{Resistance to Rotation}

\begin{figure*}[t]
	\centering{\includegraphics[scale=0.25]{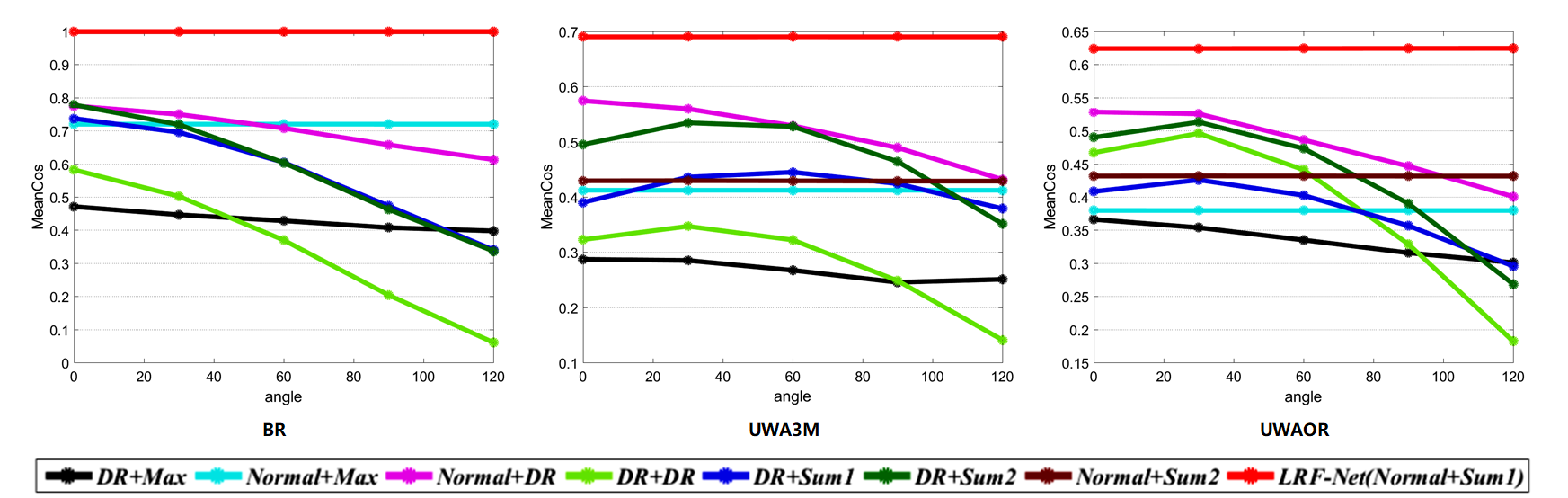}}
	\caption{Robustness performance of eight combination on three rotated datasets.}
	\label{fig_rotate}	
\end{figure*}

To evaluate the robustness of LRF-Net to rotation, we manually rotate the tested data. Specifically, we rotate the scene point clouds a certain degree among $z$-axis (i.g., 30, 60, 90, and 120 degrees). Then, we measure their $MeanCos$ performances. Fig.~\ref{fig_rotate} displays the results of eight different combinations.

As shown in Fig.~\ref{fig_rotate}, we can see that LRF-Net, Normal+Max, and Normal+Sum2 achieve very stable performances. The other ones which include "DR" part, show less robust performances. 

This result has demonstrated two conclusions. One is that it is hard to achieve rotation-invariance by only relying on original points. A guidance (e.g., normal vector) is very necessary. The other is that the invariant attributes is not indispensable. Just a simple combination (e.g., combination of original points and normal vector) can also achieve rotation-invariance. However, the invariant attributes can boost the performance
of our network.

\subsubsection{Performance under Varying Support Radius}
\begin{figure*}[t]
	\centering{\includegraphics[scale=0.25]{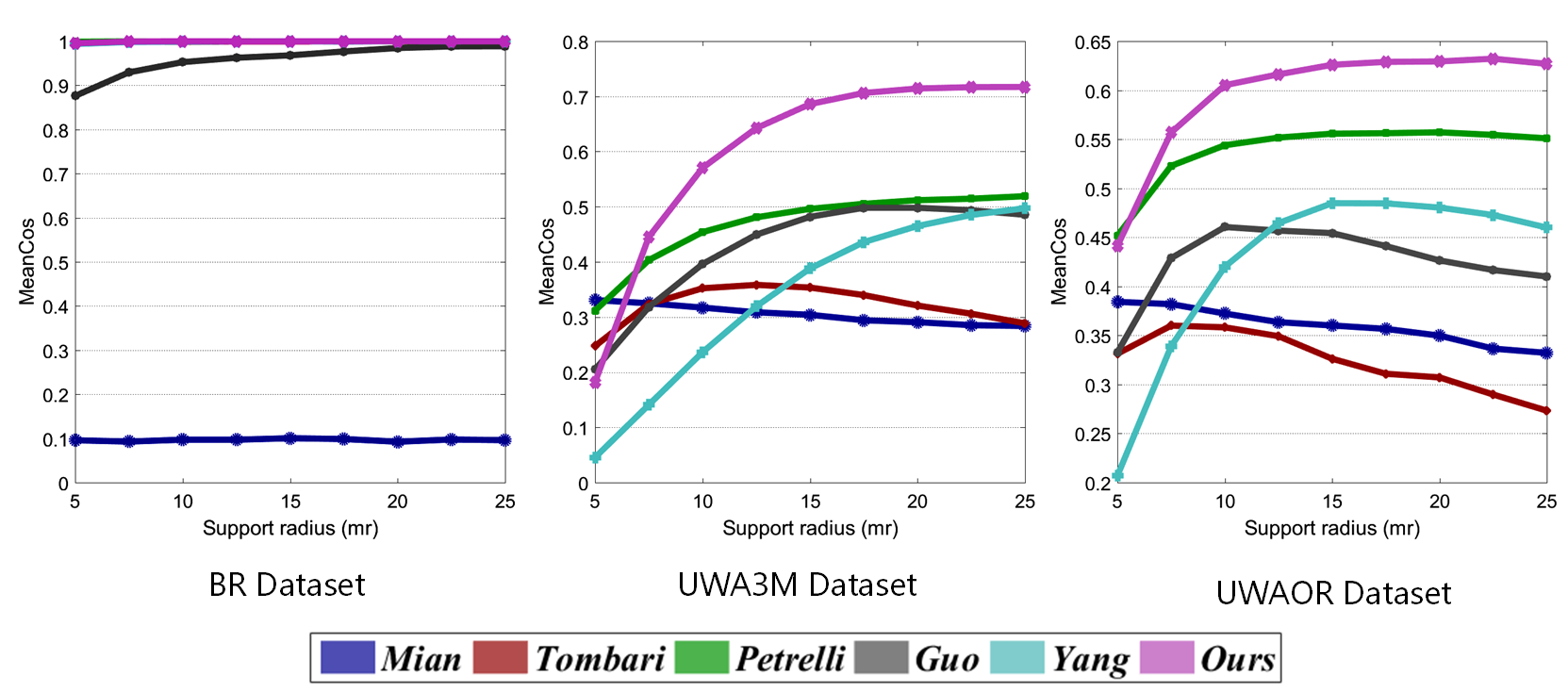}}
	\caption{$MeanCos$ performance of six LRF methods under varying support radius on three public dataset.}
	\label{fig_radius}	
\end{figure*}

Fig.~\ref{fig_radius} shows $MeanCos$ performances of six LRF methods under varying support radius on three public datasets without noise. From the observation of Fig.~\ref{fig_radius}, we can see that our LRFNet achieves a stable and outstanding performance on the BR dataset. On the UWA3M and UWAOR datasets, our LRFNet outperforms other LRF methods when support radius is more than 7.5 mr.  Another observation is that the performance of our LRFNet is tending towards stability with the increase of support radius, while some other LRF methods present a downward trend. It verifies that our LRFNet is able to gain a stable LRF from a local surface which contains enough points to guarantee its statistical significance and uniqueness.

\subsubsection{Visualization}
\begin{figure}[t]
	\centering{\includegraphics[width=0.8\linewidth]{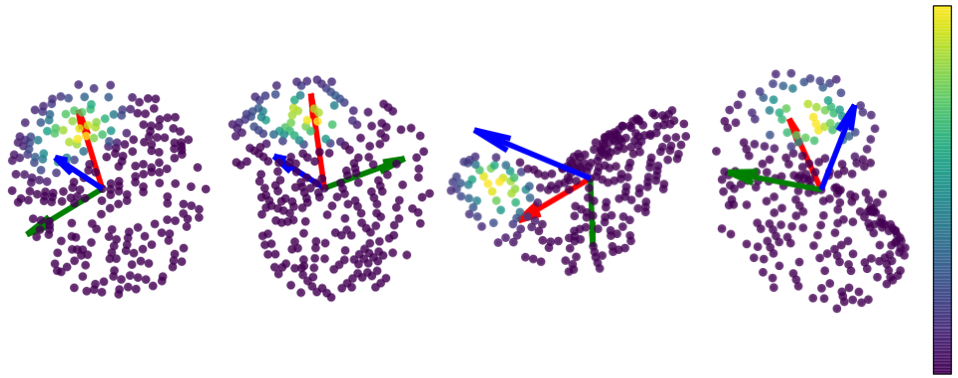}}
	\caption{The visualization of the weights for every point in a local surface.}
	\label{fig_weight}
\end{figure}

Fig.~\ref{fig_weight} visualizes the learned weights by our LRF-Net for several sample local surfaces, which presents two interesting findings. First, closer points do not seem to have greater contributions. It is a common assumption for many existing CA- and PSD-based LRF methods, including \textit{Tombari}, \textit{Guo}, and \textit{Yang}, that closer points should have greater weights. However, they are inferior to our LRF-Net in terms of repeatability performance. Second, $x$-axis estimation is generally determined by a particular area, rather than a single salient point as employed by many PSD-based methods, e.g., \textit{Petrelli}. These visualization results also demonstrate our opinion that each neighboring point in the local surface gives a unique contribution to LRF construction.

\section{Conclusion}
In this paper, we have proposed LRF-Net, a learned LRF for 3D local surface that is repeatable and robust to a number of nuisances. LRF-Net assumes that each neighboring point in the local surface gives a unique contribution to LRF construction and measure such contributions via learned weights. Experiments showed that our LRF-Net outperforms many state-of-the-art LRF methods on datasets addressing different application scenarios. In addition, LRF-Net can significantly boost the local shape description and 6-DoF pose estimation performance.  In the future, we expect further improving the LRF-Net by considering RGB cues and multi-scale geometric information.

\section*{Acknowledgment}
 This work is jointly supported by the National Natural Science Foundation of China (Grant No. U1913602)，the National Key R\&D Program of China (No.2018YFB1305504) and the Natural Science Basic Research Plan in Shaanxi Province of China (Grant No. 2020JQ-210).

\section*{References}

\bibliography{mybibfile}

\end{document}